# Glioma Classification using Multi-sequence MRI and Novel Wavelets-based Feature Fusion

**Kiranmayee Janardhan[1], Christy Bobby T[2]**

[1]Department of Electronics and Communications, Ramaiah University of Applied Sciences, Bengaluru, Karnataka, 560054, Email: kiranmayee.j@msruas.ac.in
[2]Department of Electronics and Communications, Ramaiah University of Applied Sciences, Bengaluru, Karnataka, 560054, Email: christy.ec.et@msruas.ac.in

**ABSTRACT**
Glioma, a prevalent and heterogeneous tumor originating from the glial cells can be differentiated as Low-Grade Glioma (LGG)and High-Grade Glioma (HGG) according to World Health Organization's norms. Classifying gliomas is essential for treatment protocols that depends extensively on subtype differentiation. For non-invasive glioma evaluation, Magnetic Resonance Imaging (MRI) offers vital information about the morphology and location of the tumor. The versatility of MRI allows the classification of gliomas as LGG and HGG based on their texture, perfusion, and diffusion characteristics and further for improving the diagnosis and providing tailored treatments. Nevertheless, the precise classification is complicated by tumor heterogeneity and overlapping radiomic characteristics. Thus, in this work wavelet based novel fusion algorithm were implemented on multi-sequence T1, T1-contrast enhanced (T1CE), T2 and Fluid Attenuated Inversion Recovery (FLAIR) MRI images to compute the radiomics features. Furthermore, principal component analysis is applied to reduce the feature space and XGBoost, Support Vector Machine, and Random Forest Classifier are used for the classification. The result shows that the SVM algorithm performs comparatively well with an accuracy of 90.17%, precision of 91.04%, and recall of 96.19%, F1-score of 93.53%, and AUC of 94.60% when implemented on BraTS 2018 dataset and with an accuracy of 91.34%, precision of 93.05%, recall of 96.13%, F1-score of 94.53%, and AUC of 93.71% for BraTS 2019 dataset. Thus, the proposed algorithm could be potentially implemented for the computer-aided diagnosis and grading system for gliomas.

## 1. INTRODUCTION

Gliomas, a type of tumor that occurs from the brain's glial cells, pose substantial difficulties in early detection and management due to their diverse characteristics and aggressiveness.It is essential to accurately categorize gliomas into the World Health Organization (WHO) grades I & II as LGG and grades III & IV HGG tumors to inform treatment choices and forecast patient outcomes[44]. There are two forms of aggressiveness in tumors: benign and malignant. Grades I and IIare benign, and Grade I can be removed by surgery, whereas Grade II can progress to Grade III if not removedby surgery. Grade III is highly malignant and can quickly progress to Grade IV if not removed through surgical resection. Grade IV requires both surgical resection and chemotherapy[15].Thus, the grading of glioma plays a significant role for the planning of timely and targeted treatment.

Multi-sequence datasets, like the Medical Image Computing and Computer Assisted Intervention Society (MICCAI)BraTS (Multimodal Brain Tumour Segmentation) dataset[33], are readily available, opening up new opportunities for developing sophisticated classification methods that combine machine learning and deep learning techniques. Machine learning-based glioma grading utilizing MRI data has been a popular study area.

MRI is a critical component in the classification and assessment of gliomas, a challenging subtype of brain tumors. This approach heavily relies on the use of different MRI sequences such as gadolinium-enhanced T1-weighted imaging, T2-weighted imaging, T1-weighted imaging and Fluid Attenuated Inversion Recovery (FLAIR). The location, size, shape, and characteristics of the tumor—such as its degree of enhancement, diffusion patterns, and perfusion patterns—are indicated by these sequences in significant detail. The ability to differentiate between many glioma subtypes and grade the aggressiveness of each allows for the formulation of individualized treatment plans, ultimately improving diagnostic precision and the standard of care provided to patients [35].

Wavelets are flexible mathematical operations used in data analysis and signal processing. It captures high and low-frequency components with significant time and frequency localization. Applications include denoising, feature extraction, pattern recognition, and picture and audio compression[28]. As the wavelets effectively represent and manipulate images at various scales, it is especially favored for image processing applications. A significant number of quantitative aspects from medical images, frequently derived from radiological scans like MRI, Computed Tomography (CT), or Positron Emission Tomography (PET), are extracted and analysed as part of the study area known as radiomics[2, 6, 18, 22, 32, 39, 41]. These features capture the glioma's shape, texture, intensity, and spatial relationships. Due to its capacity to record spatial relationships and quantify texture features using statistical measures, including contrast, homogeneity, energy, and correlation, the Gray-Level Co-occurrence Matrix (GLCM) is frequently used in texture analysis for feature extraction. It is robust to changes in image orientation since it is insensitive to rotation and translation. The appeal of GLCM in medical imaging can be attributed to its computational effectiveness, simplicity of implementation, and capacity to record discriminative texture patterns[16]. Ultimately, GLCM is a flexible and successful method for identifying complex textures in images, making it possible for varied applications to perform classification, segmentation, and pattern recognition[38].

The shape features are the geometrical characteristics of regions in animage. Volume, surface area, compactness, and sphericity are typical shape characteristics. It provides information on the shape and size of gliomas. The Region of Interests (ROIs) pixels' intensity levels used to compute the first-order features. It measures mean, median, variance, skewness, and kurtosis along with the distribution of pixel intensities. The GLCM features[6, 22]use second-order statistics for features that can infer the degree of correlation between pairs of pixels. Contrast, energy, homogeneity, and correlation are the derived properties. These attributes record the image's textural patterns and variances. The Gray-Level Dependence Matrix (GLDM)[39]features are based on the gray-level intensity values of the pixels, GLDM quantifies the dependence between pairs of pixels. It has characteristics like homogeneity, correlation, and contrast that shed light on the complexity and variability of textures. The Gray-Level Run Length Matrix (GLRLM) [18]featuresfocus on measuring the distances between adjacent pixels that are similar in intensity along various axes. Short-run and long-run emphasis are features that can capture the length distribution of similar intensity values. The Gray-Level Size Zone Matrix (GLSZM) [41]features describe the distribution of linked sections in an image with a specific gray-level value and size. It reveals details on how these regions are arranged spatially. The Neighboring Gray Tone Difference Matrix (NGTDM) [2]features analyse the variations in gray tones between a pixel and its neighbors. Contrast, coarseness, and busyness are characteristics that represent regional texture variations.

Principal Component Analysis(PCA), is used in data analysis and machine learning to minimize dimensionality while retaining the most critical data[37].It captures large variations by orthogonalizing characteristics into parts. To minimize the dimensionality of the data, selecting just a portion of these principal components is performed[19]. It helps with visualization and enhances model performance by reducing redundancy, making it useful for high-dimensional datasets. Machine learning classifiers are algorithms that divide data into predetermined classes in supervised learning[42]. They are essential in numerous applications, including image recognition, spam detection, and medical diagnosis. It is a usual practice to test various classifiers and compare their performances. Machine learning classifiers have changed data analysis and decision-making in many different sectors. The best classifier should be chosen based on the dataset's particular characteristics, the problem's complexity, the need for interpretability, and the available computational resources. In this work, the best classifier has been found by experimenting with XGBoost, SVC, and RFC and comparing their results. With the help of these classifiers, numerous fields' data analysis and decision-making processes have evolved dramatically, enabling precise predictions and insights from various datasets[40].

This paper is organised as follows. Section 2 contributes state-of-the-art advancements, as well as contemporary baseline methodologies, to aid in the development of a decision support system as a research contribution. Section 3discusses the details of datasets used i.e., BraTS 2018 and 2019, pre-processing along with different ROIs considered. It is followed by the proposed technique for fusion of the wavelet transform to generate the fused images along with feature extraction methods used which are radiomics based techniques. This is followed by features reduction technique in detail. Moreover, this section presents the exhaustive experimentation and results over BraTS 2018 and 2019 datasets considering feature selection, reduction, and extraction techniques along with our proposed model demonstrating the effectiveness of our approach with its implementation on training and testing the classifiers. This is followed by the evaluation metrics. Section 4provides the results and discussion and Section 5, concludes the work along with future scope.

## 2. Related Works

In [11], the authors developed machine learning models for differentiating LGGand HGG were performed using normalized multi-parametric Magnetic Resonance Imaging (mp-MRI) features. To choose features for the model, the SVM-based recursive feature elimination method was used when implemented on a local dataset; the accuracy obtained was 93%, and its drawback is that it was trained on a limited patient population. In another study [14], the researchers use radiomic characteristics that are in line with World Health Organization (WHO) standards to examine how the LGG population is processed using SVM classifier on the BraTS dataset. The LGG and HGG data are used to train the SVM classifier, and its performance is assessed by examining its classification parameters. Using this method, the researchers evaluated the SVM classifier's ability to differentiate between LGG and HGG images of BraTS dataset. The resulting accuracy was 84.1%, where the model contained false positives.

The authors in[13],worked on all four multi-sequence MRIs to extract radiomic characteristics, including T1, contrast-enhanced T1, T2, and FLAIR. The Least Absolute Shrinkage Selection Operator (LASSO), the two-sample T-test, and a feature correlation threshold were all used in the feature selection procedure. This procedure was used onindividual MRIs in various combinations. After that, anRFC was employed to distinguish between HGG and LGG using the chosen features. This method efficiently classifies and distinguishes between HGG and LGG based on multi-parametric MRIs by utilizing radiomic characteristics and the RF algorithm. The accuracy obtained was 91.3% on BraTS 2015. Previous research has emphasized the importance of patient age in feature extraction comparisons between two groups of glioma patients—those under and over 45 years. In the study[10], the authors worked on the effectiveness of the glioma-grade classification models and assessed them using a five-fold cross-validation. The algorithm known as Minimum Redundancy Maximum Relevance was used to choose instructive features from the training set. Following that, classification utilizing the chosen features was applied to RF, SVM, and Logistic Regression (LR) techniques. The results were obtained in terms of Area Under the Curve(AUC). The AUC for logistic regression is 90.1%, SVM 88.6%, and RF 92.1%.

In the research work[43], the authors gave a quantitative interpretation and pinpointed the key factors that affect glioma grading, the Local Interpretable Model-Agnostic Explanations (LIME) algorithm was used. This method offers insights into the grading model's decision-making process and highlights the significance of criteria in the classification process. The accuracy is 90% when implemented on a local dataset. In the study [25], the authors classified gliomas (GradeII–IV), using a CNN-based approach in conjunction with a Genetic Algorithm (GA) was implemented. The CNN model carried out the grading task, and the GA optimized the CNN model's parameters. By combining deep learning and genetic optimization techniques, this integrated strategy seeks to improve the precision and efficiency of glioma grading, which obtained an accuracy of 90.9% on the REpository for Molecular BRAin Neoplasia DaTa (REMBRANDT) dataset[20].

The authors in[28] enabled a thorough examination of the texture and spatial aspects of the images by using wavelet analysis to extract pertinent features from MRI. The accuracy obtained was 97.62% on the BraTS 2018 dataset, but the computing time was high. In the study[1], the authors worked onthe Weighted Neighbour Distance using the Compound Hierarchy of Algorithms Representing Morphology (WNDCHRM) tool-based classifier, and the VGG-19 Deep convolutional Neural Network (DNN) as the two classifiers in the research for analysis. When implemented on a local database, the accuracy was 92.86% for WNDCHRM approach and 94.64% for DNN. In the work[21], the authors used learned features derived from a trained Convolutional Neural Network (CNN) for the goal of glioma grade prediction. These newly acquired elements from CNN's deep learning method increased the study's analysis' accuracy and dependability of glioma grade predictions. The accuracy is 87% on a local dataset. Although this study did not consider molecular information, replicating the findings using a large dataset would be auseful addition.

The study[38], the authors included a variety of imaging features and clinical measurements, including both basic texture features (such as intensity and morphology) and sophisticated texture features (such as a GLCM and GLRLM). These features were taken from the sub-images and used to train a linear SVM model. The study used this strategy to better use a broad range of features for classification and analysis inside the SVM framework. The accuracy is 75.12% when implemented on a local dataset. The dataset employed in this study is small and narrowly focused, and it has yet to be thoroughly tested on industry-standard datasets.

## 3. METHODOLOGY

### 3.1 Datasets & Image Pre-Processing

The operational workflow of this work for the classification of gliomas into HGG/LGG is shown in Figure 1. The BraTS' 2018 and 2019 MRI datasets are used for this study [33]. These datasets include FLAIR, T1,

T1CE, and T2 images from multisequence MRI scans. Table 1 shows the number of HGG and LGG images and their corresponding segmentation masks that are used in this analysis. The size of the images is240X240X155 in the Neuro imaging Informatics Technology Initiative (NIfTI) file format [31]. These images cannot be used straight away to train models;thus, the images were transformed into NumPy arrays with 128x128 dimensions. This conversion operation aidsin lowering the computational cost. Further, the NumPy array is normalized to ensure the uniformity of input data.

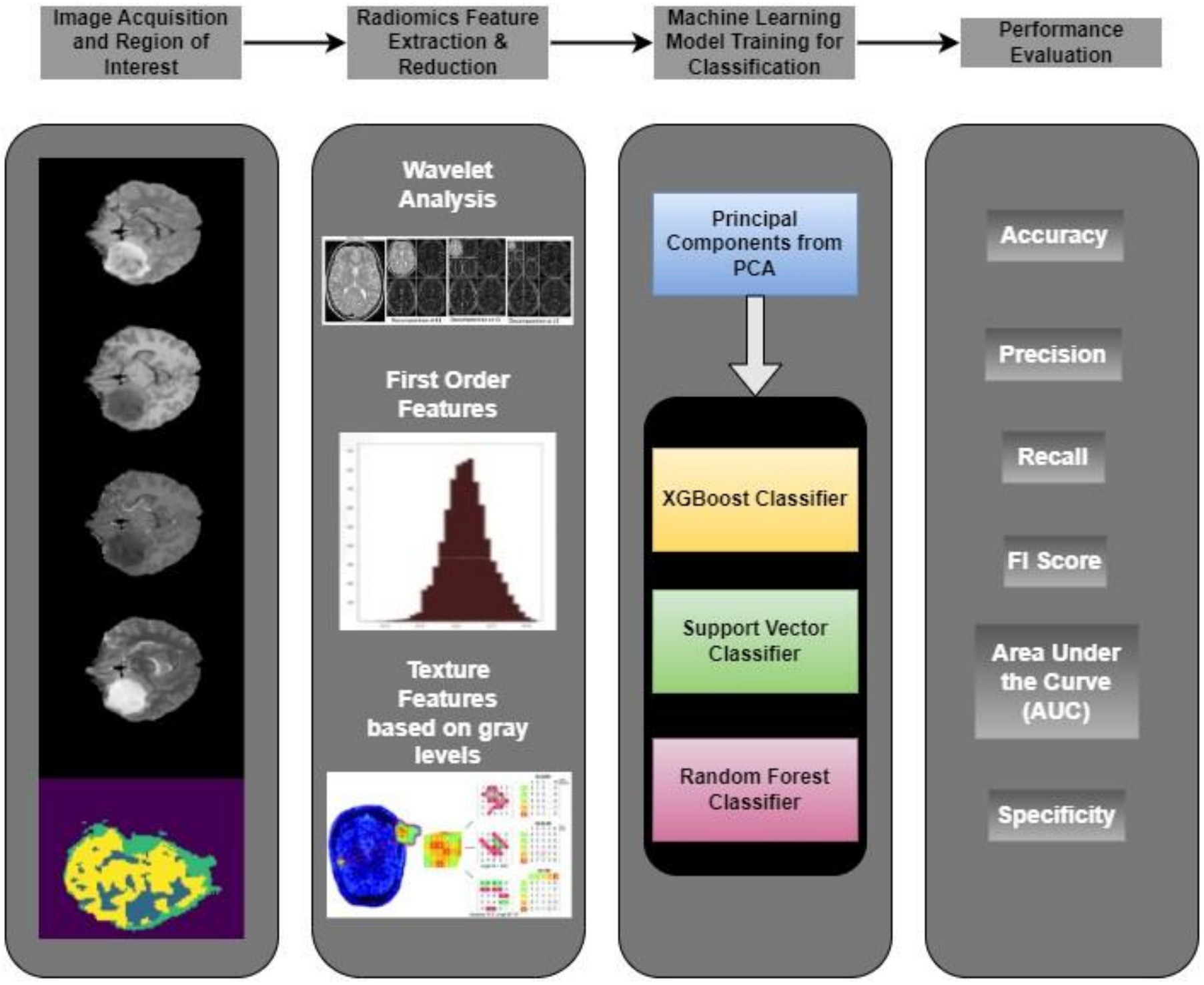

**Figure 1:** Work flow of glioma classification using Wavelet based multi-sequence MRI fusion

**Table 1:** BraTS Datasets showing the number of cases for HGG/LGG.

| Datasets | HGG | LGG |
|---|---|---|
| BraTS 2018 | 210 | 75 |
| BraTS 2019 | 259 | 76 |

### 3.2 Regions of Interest

Different Regions of Interest (ROIs) in the analysis of gliomas are essential for comprehending the nature and development of the tumor. These ROIs include necrosis, which represents the tumor's non-functional core; edema, which depicts the area around the tumor where swelling and inflammation have developed. The enhancing region represents the portion of the tumor that is actively growing; and the non-enhancing region contains infiltrative tumor cells. Each of these ROIs offers valuable insight into the behavior of the tumor, assisting doctors in customizing treatment plans, gauging therapeutic outcomes, and making decisions that will lead to better patient care. [46]

In this work, instead of concentrating primarily on one particular tissue type for the identification of glioma, the goal was to collect the textural features from the various tissue types of the tumor. The three ROIs representing tissue types that are considered to extract radiomics features are shown in Figure 2.The regions, ROI1 indicates the No Tumor Region, ROI2 indicates the Necrotic and Non Enhancing Tumor Region and ROI3 indicates the Tumor Core Region and the Peritumoral Edema which are considered for classification. TheROI2 is considered for the classification of LGG and ROI3 which represents the whole tumor is considered for the classification of HGG as they are typically characterized by contrast-enhanced areas on T1-weighted imaging and hype rintensity on FLAIR [7].

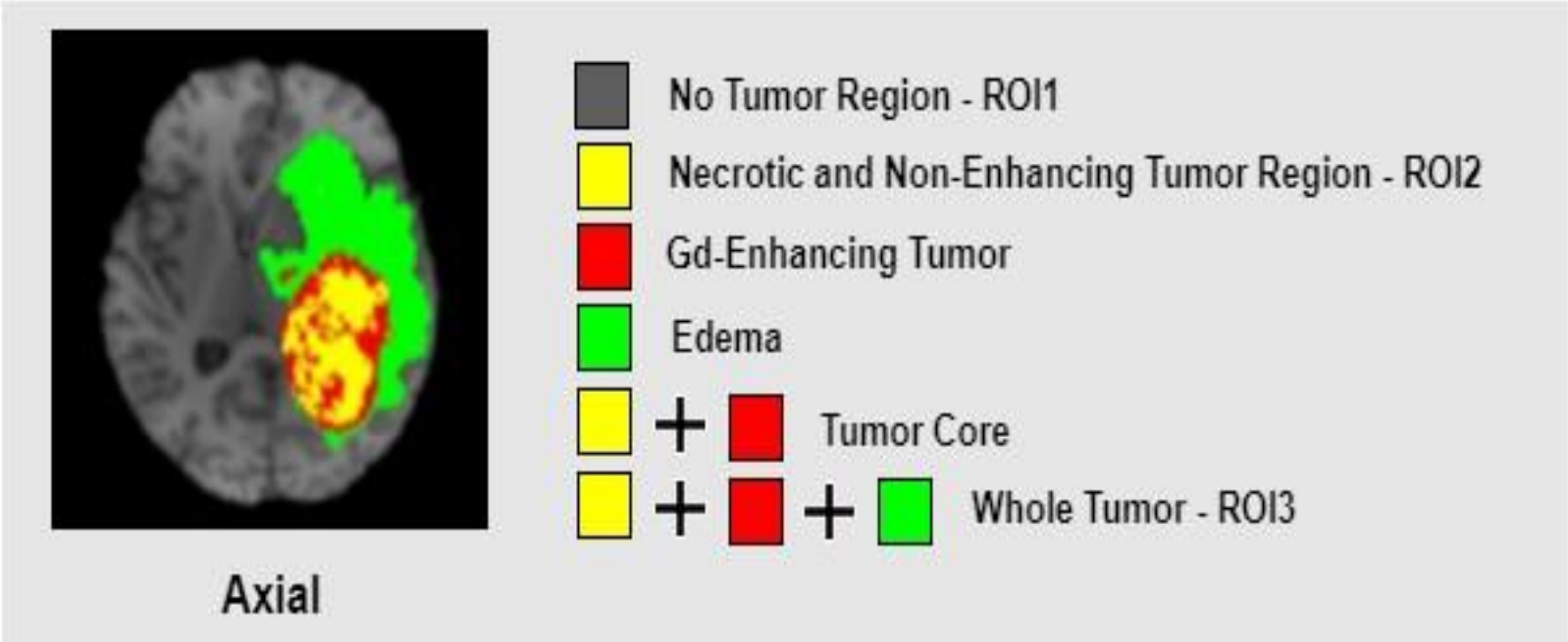

**Figure 2:** Axial plane MRI scan indicating different regions of interest [5].

### 3.3 Fusion of MRI Sequences using Discrete Wavelet Transform

In this model, Discrete Wavelet Transform (DWT) was used for efficient textural feature extraction of the images. The wavelet coefficients are calculated at discrete scales based on powers of 2, using the equation:

$$W_{j,\ k}(n) = \sum_{j}\sum_{k} x(k) 2^{-j/2}\, \Psi\, (2^{-j}\, n - k)$$

The discrete function $(k)$ represents the detailed component which is the weighted summation of wavelets plus a coarse approximation. $\Psi(\,)$ is the wavelet function for variables $j$ and $k$,which serve as the scaling and translation parameters, and$n$ represents the discrete position. The coarse approximation was further decomposed by using iterative low-pass and high-pass filtering.

The approximation and detailed coefficients at scale $(j+1)$ at position $k$ are calculated by convolving the current approximation coefficients $a_j[m]$ with $l[m]$ and $d_j[m]$with$\hbar[m]$, translated by $(-2k)$, respectively using the equations as follows:

$$a_{j+1}[k] = \sum_{m=-\infty}^{+\infty} l\,[m - 2k]\, a_j\,[m]$$

$$d_{j+1}[k] = \sum_{m=-\infty}^{+\infty} \hbar\,[m - 2k]\, a_j\,[m]$$

The sequences $l\ [m\ -2k]$ and $\hbar\ [m\ -2k]$ are low-pass and high-pass components.$a_j[m]$is the approximation coefficient at scale $j$ which captured the low-frequency components and $d_j[m]$is the detailed coefficient at scale $j$ which captured the high-frequency components of the image.The four sub-frequency bands generated are Horizontal Detail (HD), Vertical Detail (VD), Diagonal Detail (DD), and Approximate Detail (AD) coefficients.

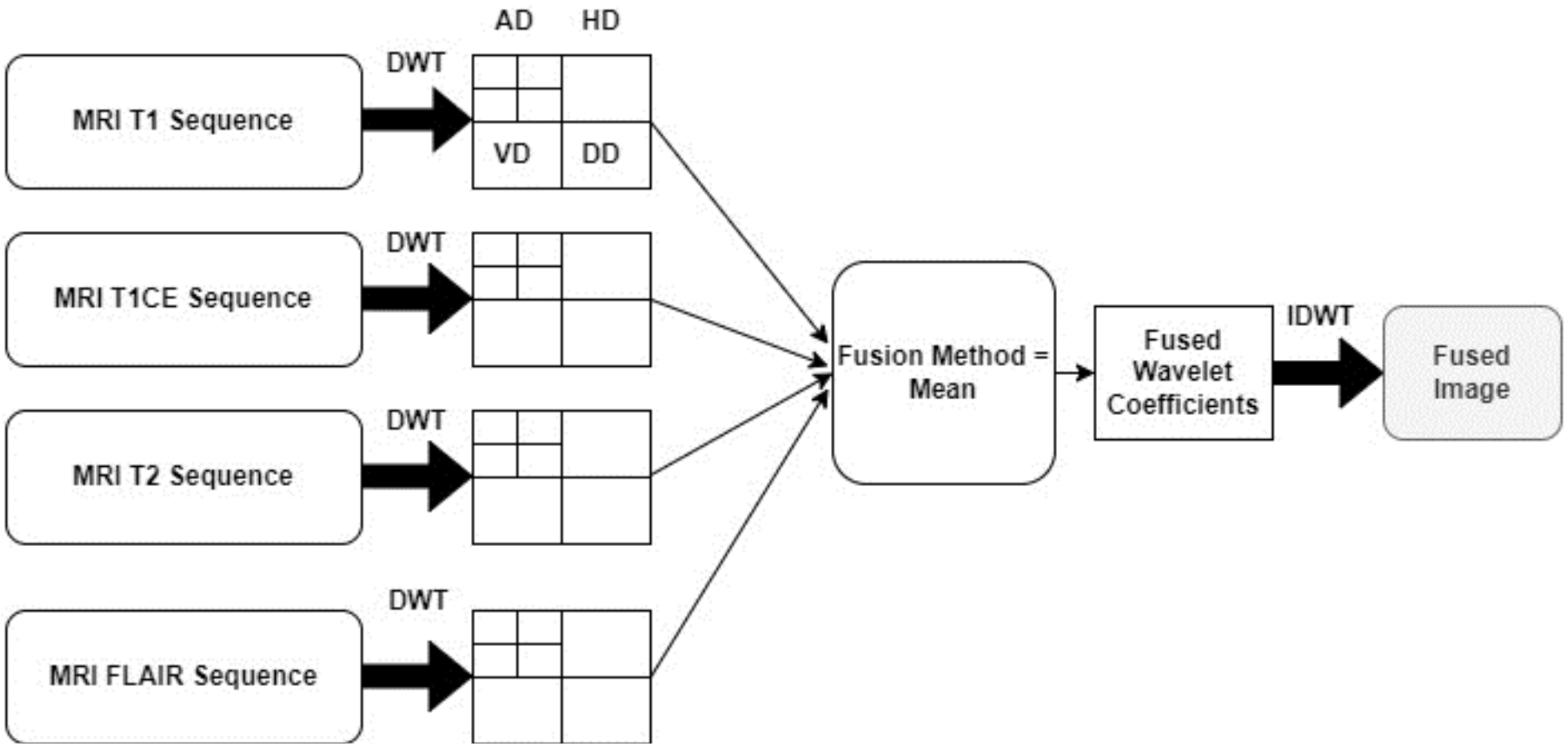

**Figure 3:** Representative block diagram of novel wavelet fusion technique

Figure 3 shows the novel wavelet fusion technique implemented in this work for fusing FLAIR, T1, T1CE, and T2 images for the extraction of high-quality textural features.Wavelets were chosen to give high-quality featuresIn this methodology, Daubechies wavelet db1 was applied on all these four sequence images to generate wavelet coefficients, and these derived coefficients are fused using the mean value computation technique[26] to geta fused image. The fused image is generated by applying Inverse

Discrete Wavelet Transform (IDWT) and normalized to be in the range of 0 to 255 to get the gray scale image. The complete set of fused images obtained is further super imposed on each of their ROIs respectively which are utilized in the extraction of the radiomics features. This superimposition of the ROIs ensures that feature values pertaining only to those regions are generated.

### 3.4 Feature Extraction with Radiomics

The radiomics features are powerful tools for assessing minute textural variations in medical imaging that might not be visible to the human eye. It supports more precise diagnosis, evaluation of the treatment's effectiveness, and glioma prognosis in various medical imaging modalities, including MRI. In this work, the radiomics features such asshape[6, 32], first order [22], GLCM, GLDM, GLRLM, GLSZM,and NGTDMare computed on three different ROIs (ROI1, ROI2, and ROI3)of the superimposed images. Table 2 shows thelist of radiomics features [47] extracted from the superimposed image using the Radiomics Feature Extractor. The number of features extracted was 107 for each ROI. This aggregation ensures that relevant information from the multiple MR imaging sequences and the various segmented tumor regions are considered for the classification process. These extracted features are stored in separate Panda'sdata frames for each ROI. The column names of the data frames are modified to include the ROI number and tumor grade for easy identification. The data frames for each ROI are merged into a single data frame and the features were saved to LGG and HGG feature files with sizes 277 KB and 754 KB for BraTS 2018 and 281 KB and 926 KB for BraTS 2019respectively.

### 3.5 Dimensionality Reduction with PCA

Principal Component Analysis (PCA) helps reduce the feature space while retaining the most essential information. The selection of the principal components explains a significant amount of variance in the data. This has been achieved by transforming a new set of variables called the Principal Components (PCs). PCs are uncorrelated and ordered, and the first few retain most of the variation in the original variables. The Eigende composition of the covariance matrix $(X^T X)$ is given as:

$$Score\colon\ T = X\,W$$

where each column W is a Principal Component, $W_1, W_2 \ldots. W_n$ and $W_1$ account for more data variance than $W_2$. The principal components in $1^{st} r$ columns in the new dimensional space are used for further analysis. The variations included in the data inherently influence the PCs. The errors caused by dimension reduction can be minimized by selecting the PCs that significantly contribute to the overall variance [24]. In this work, PCA was applied to the features and the results yielded two principal components while preserving over 85% of the total data variance that is used in the classification of LGG and HGG images[23].

### 3.6 Training the Classification Models

In this work, the training and evaluation of models have been performed using eXtreme gradient boosting (XGBoost) [34], Support Vector Classifier (SVC) [12], and Random Forest Classifier (RFC)[9] to distinguish between HGG and LGG images. XGBoost is an efficient and powerful gradient boosting algorithm widely used in machine learning for both classification and regression tasks. SVC is a supervised machine learning algorithm that finds an optimal hyperplane to separate data into different classes, maximizing the margin between classes while relying on support vectors for decision boundaries. RFC is an ensemble learning method that combines multiple decision trees to improve predictive accuracy and reduce overfitting by averaging or voting on the individual trees' predictions.

XGBoost is a highly powerful scalable tree-based machine learning framework that combines multiple weak classifiers (decision trees) to create a stronger model. The final prediction is obtained by summing the predictions from all the individual trees, weighted by their importance, and is given by the equation as follows:

$$y_p = \sum \left(T(x,\ \theta_t)\right) + b$$

Where $y_p$ is the predicted output, $\left(T\left(x,\ \theta_t\right)\right)$ is the prediction from the decision tree with parameters $\theta_t$, and $b$ is the bias term. A 1000 decision trees were built during the training process as it regulated the overall complexity and quantity of iterations for the boosting process. The random state was set to 0 to ensure that the algorithm yields the same outcomes every single time it is implemented with the same data. The objective parameter was set to logistic for binary classification.

**Table 2:** List of radiomics texture features extracted from fused MRI images[32]

| SHAPE | FIRST ORDER | GLCM | GLDM | GLRLM | GLSZM | NGTDM |
|---|---|---|---|---|---|---|
| • Elongation<br>• Flatness<br>• Least Axis Length<br>• Major Axis Length<br>• Maximum 2D<br>• Maximum 3D<br>• Mesh Volume<br>• Minor Axis Length<br>• Sphericity<br>• Surface Area<br>• Surface Volume<br>• Voxel Volume | • 10Percentile<br>• 90Percentile<br>• Energy<br>• Entropy<br>• Interquartile Range<br>• Kurtosis<br>• Maximum<br>• Mean Absolute Deviation<br>• Mean<br>• Median<br>• Minimum<br>• Range<br>• Robust Mean Absolute Deviation<br>• Root Mean Squared<br>• Skewness<br>• Total Energy<br>• Uniformity<br>• Variance | • Autocorrelation<br>• Cluster Prominence<br>• Cluster Shade<br>• Cluster Tendency<br>• Contrast<br>• Correlation<br>• Difference Average<br>• Difference Entropy<br>• Difference Variance<br>• Inverse Difference<br>• Inverse Difference Moment<br>• Inverse Difference Moment Normalized<br>• Inverse Difference Normalized<br>• Informational Measure of Correlation1<br>• Informational Measure of Correlation2<br>• Inverse Variance<br>• Joint Average<br>• Joint Energy<br>• Joint Entropy<br>• Maximal Correlation Coefficient<br>• Maximum Probability<br>• Sum Average<br>• Sum Entropy<br>• Sum Squares | • Dependence Entropy<br>• Dependence Non-Uniformity<br>• Dependence Non-Uniformity Normalized<br>• Dependence Variance<br>• Gray Level Non-Uniformity<br>• Gray Level Variance<br>• High Gray Level Emphasis<br>• Large Dependence Emphasis<br>• Large Dependence High Gray Level Emphasis<br>• Large Dependence Low Gray Level Emphasis<br>• Low Gray Level Emphasis<br>• Small Dependence Emphasis<br>• Small Dependence High Gray Level Emphasis<br>• Small Dependence Low Gray Level Emphasis | • Gray Level Non-Uniformity<br>• Gray Level Non-Uniformity Normalized<br>• Gray Level Variance<br>• High Gray Level Run Emphasis<br>• Long Run Emphasis<br>• Long Run High Gray Level Emphasis<br>• Long Run Low Gray Level Emphasis<br>• Low Gray Level Run Emphasis<br>• Run Entropy<br>• Run Length Non-Uniformity<br>• Run Length Non-Uniformity Normalized<br>• Run Percentage<br>• Run Variance<br>• Short Run Emphasis<br>• Short Run High Gray Level Emphasis<br>• Short Run Low Gray Level Emphasis | • Gray Level Non-Uniformity<br>• Gray Level Non-Uniformity Normalized<br>• Gray Level Variance<br>• High Gray Level Zone Emphasis<br>• Large Area Emphasis<br>• Large Area High Gray Level Emphasis<br>• Large Area Low Gray Level Emphasis<br>• Low Gray Level Zone Emphasis<br>• Size Zone Non-Uniformity<br>• Size Zone Non-Uniformity Normalized<br>• Small Area Emphasis<br>• Small Area High Gray Level Emphasis<br>• Small Area Low Gray Level Emphasis<br>• Zone Entropy<br>• Zone Percentage<br>• Zone Variance | • Busyness<br>• Coarseness<br>• Complexity<br>• Contrast<br>• Strength |

Logloss was implemented as the evaluation metric which stands for logarithmic loss and measures how well the predicted probabilities align with the actual classification of LGG/HGG [33].

Support Vector Classifier finds an optimal hyperplane in high-dimensional space to separate different classes and seeks to maximize the margin between the classes while minimizing the misclassification errors. It is given as:

$$y_p = sign\left(\sum (y_i * a_i * K(x_i, x)) + b\right)$$

Where $y_p$ is the predicted output, $y_i$ is the class label of the $i^{th}$ support vector, $a_i$ is the Lagrange multiplier, $x_i$ is the $i^{th}$ support vector, $K(x_i, x)$ is the kernel function that measures the similarity between $x_i$ and input $x$, and $b$ is the bias term. In this work, the regularization parameter (cost) was set to 1 for having a trade-off between maximizing the margin of the decision boundary and correctly classifying training the HGG and LGG instances. The kernel was set to radial basis function [45] and the degree of the polynomial kernel was set to 3 to control the adaptability of the decision boundary. The parameter gamma is used to supervise the shape of the decision boundary. It was set as follows:

$$gamma = \frac{1}{numberoffeatures * variance}$$

The shrinking parameter was set to Boolean True which can speed up training for datasets. The tolerance for stopping the optimization process was to set $1e^{-1}$. The probability estimations for classification labels were enabled by setting the probability parameter to True. In addition to making predictions, the SVC was trained to produce probability estimates for each class[27].

Random Forest Classifier is an ensemble learning method that constructs multiple decision trees and combines their predictions through voting or averaging. Each decision tree is trained on a random subset of the data with replacement, and the final prediction is obtained by aggregating the predictions from all the trees. It is given by the equation:

$$y_p = majority_{vote}(T_1(x), T_2(x), \ldots, T_n(x))$$

Where $y_p$ is the predicted output, $majority_{vote}()$ function is the majority vote mechanism, $T_i(x)$ are the individual decision trees and input $x$. $(T_1(x), T_2(x), \ldots, T_n(x))$ work together to make a collective decision. In this work, the random seed number was set to 0 to ensure that the algorithm's performance was consistent. The number of decision trees or estimators was set to 1000 in the random forest to ensure the improved behavior of the classifier [4, 17]. The quality of the split in the decision trees was set to Gini which computes the impurity that is designed to decrease the probability of incorrect classification of a sample [17]. The maximum features that can be taken into consideration while dividing each tree node were assigned to the square root of the total number of characteristics. The minimum number of samples needed to divide an internal node in a decision tree is set to 2. A higher value can avoid overfitting, but if set too high, it could result in underfitting. The minimum of samples present in a leaf node is set to 1 and by adjusting this value, overfitting is prevented, and the size of the tree is managed.

### 3.7 K-fold Cross-Validation

The model is trained and tested K times, with each evaluation utilizing the remaining (K-1) folds as the training set and a different fold as the validation set. The average performance results from all K-iterations make up the final performance metric. This cross-validation offers a more thorough evaluation of a model's performance on unobserved data by repeatedly training and testing the model on multiple subsets of the data. This helps to estimate a model's generalization performance better. This study used K-fold cross-validation[3, 10], where K=5, to segregate the training and test cohorts because it was a single source study (it only used data from the BraTS database), which helped to reduce overfitting. In each fold, the ratio of HGG and LGG was maintained between the training dataset and test dataset.

### 3.8 Evaluation Metrics

The efficacy of glioma classification algorithms has been assessed using quantitative classification metrics[36]such as Accuracy, Precision, Recall/Sensitivity, F1-score, Specificity, and Confusion Matrix. Accuracy gauges how well the classification was made overall as it measures the proportion of correctly identified samples to all samples. Accuracy gives a preliminary impression of how well the algorithm works, but it may be deceptive if the classes are imbalanced. Precision is the fraction of correctly predicted positive outcomes measured and is also known as positive predictive value. In the context of glioma categorization, precision is used to identify the proportion of predicted glioma cases that were indeed true gliomas. Low false positive rates are indicative of high precision. Recall is the fraction of real positive cases that were accurate,and it demonstrates how successfully the algorithm distinguishes between actual gliomas in glioma classification. The F1-Score is the low false negative rate and is indicated by a high recall value. The harmonic mean of recall and precision is known as the F1-score. It offers a fair assessment that considers both false positives and false negatives and is beneficial in imbalanced classes.Specificity is the fraction of actual negative cases correctly predicted and is also known as the true negative rate. It explains how successfully the algorithm distinguishes non-glioma cases in the context of glioma classification. These quantitative metrics clarify how effectively each algorithm works to classify gliomas.

The confusion matrix[36] tabulates the predicted and actual class labels for a dataset to represent the performance visually. Figure 4 shows the Four groups into which the results are categorized True Positive (TP), False Positive (FP), True Negative (TN), and False Negative (FN). Positive classes are denoted as the HGG cases, while negative classes are categorized as the LGG cases. TP is representative of the number of HGG cases that the model correctly predicted. TN represents the number of LGG cases that are correctly identified by the model. This matrix makes it possible to grasp a model's accuracy, precision, recall, and F1 score, evaluating its efficacy for the two classes and directing prospective enhancements to improve classification performance. It is a key tool for evaluating the effectiveness of classification algorithms. The confusion matrix illustrated a variety of insights in this work, such as class-specific performance. This feature enabled the evaluation of how well the classifiers performed for each class separately and which classes the model predicts well and which it does poorly.

| | | Predicted Class | |
|---|---|---|---|
| | | HGG | LGG |
| Actual Class | HGG | True Positive | False Negative |
| | LGG | False Positve | True Negative |

**Figure 4:** Confusion matrix for the glioma classification

The Receiver Operating Characteristic(ROC) curves[10] plot the True Positive Rate (TPR) against the False Positive Rate (FPR) as the discrimination threshold varies. The diagonal line in the ROC space represents the performance of the classifier. The Area quantifies the classifier's total discriminative power Under the Curve (AUC). A high AUC demonstrates an excellent capacity to discriminate between HGG and LGG cases. ROC curves enabled performance comparison by comparing the AUC values of these classifiers.

## 4. RESULTS & DISCUSSION

In this work, the wavelets-based fusion of MRI scans has been implemented to generate distinctive radiomics features for both HGG and LGG. The Figure 5 and 6 shows the representative images of FLAIR, T1, T1CE, T2, segmentation mask and the wavelet fused images of LGG and HGG respectively for BraTS 2018 and 2019 datasets.

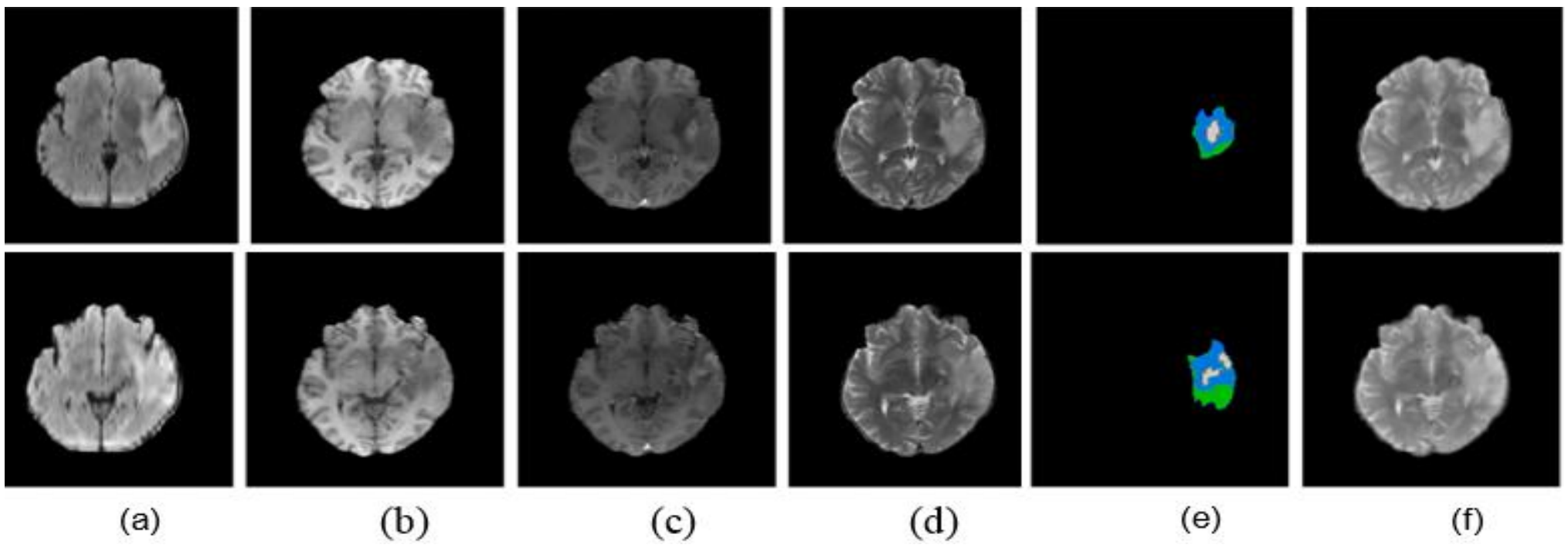

**Figure 5**:(a) FLAIR (b) T1 (c) T1CE (d) T2 (e) Segmentation mask (f) Wavelet fused images of LGG from BraTS 2018 and BraTS 2019 datasets

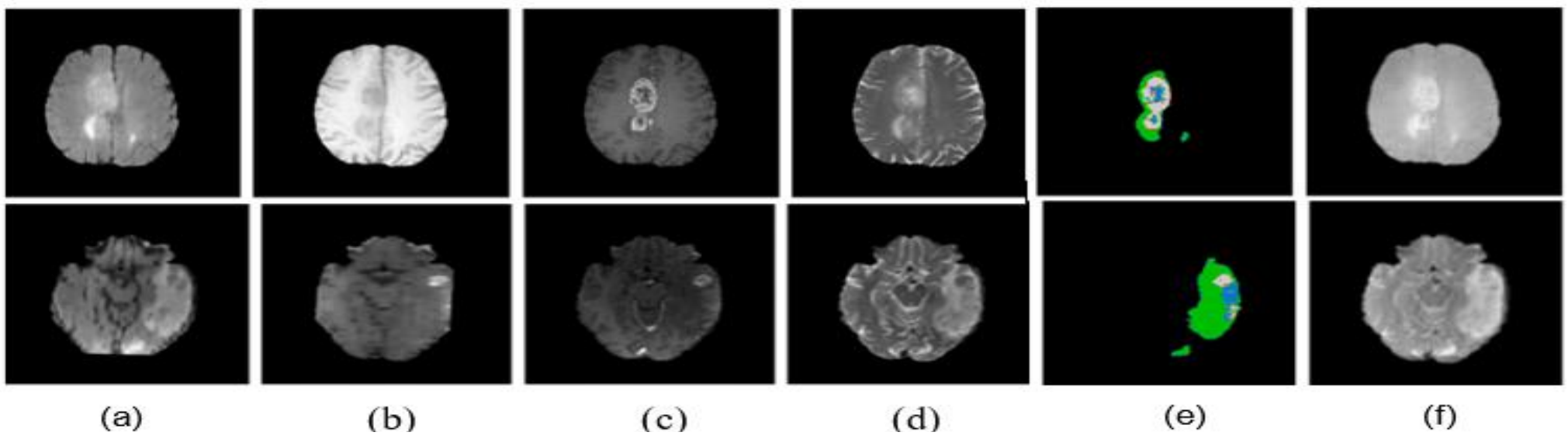

**Figure 6:** (a) FLAIR (b) T1 (c) T1CE (d) T2 (e) Segmentation mask (f) Wavelet fused images of LGG from BraTS 2018 and BraTS 2019 datasets

The loading values of Energy ROI1 and Energy ROI2 for LGG and HGG are as shown in Table 3 respectively. These high loading values represent the contribution of a principal feature for further processing.

**Table 3:** Loading Values of the first two principal components

| Principal Components | Loading Value | | | |
|---|---|---|---|---|
| | **2018** | | **2019** | |
| | **LGG** | **HGG** | **LGG** | **HGG** |
| Energy ROI2 | 0.545671668 | 0.6194966 | 0.546236746 | 0.623383504 |
| Energy ROI1 | 0.324592126 | 0.3067342 | 0.324757124 | 0.300574878 |

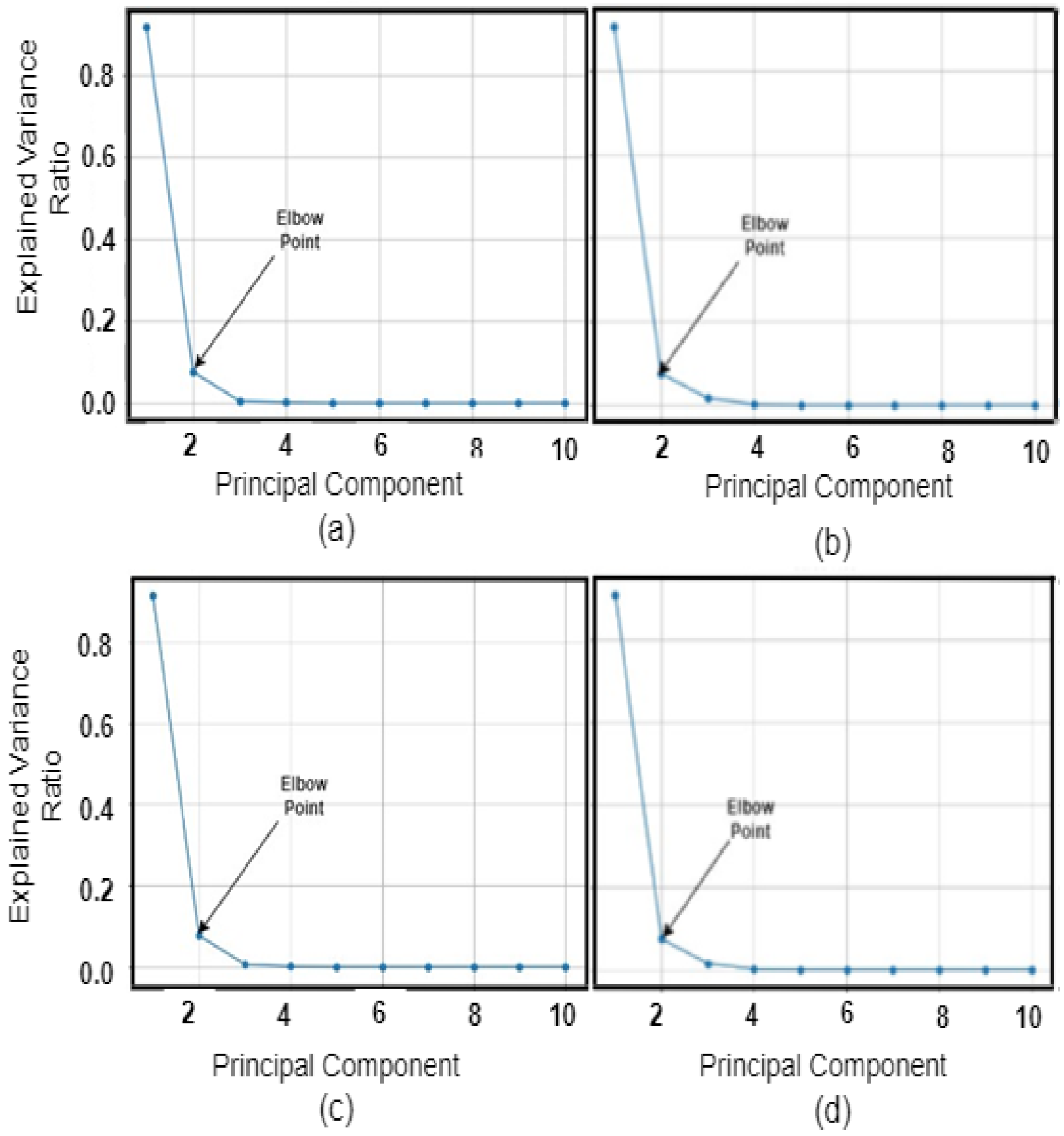

**Figure 7:** The scree plot representing the PCs for (a) LGG -BraTS 2018 (b) HGG - BraTS 2018 (c) LGG - BraTS 2019 and (d) HGG - BraTS 2019

Figure 7 shows the scree plots with explained variance ratio versus the principal components for HGG and LGG features. With an increasing number of principal components, the amount of explained variance typically gradually increases, as additional components capture more information from the data, although each additional component explains less variance compared to the earlier ones. The elbow point can be seen at PC2 where the slope becomes less steep. This is seen in both the plots for LGG and HGG. The highest amount of variance in the data is captured by PC1. This indicates that it explains the majority of the variance among the variables. Even while PC2 captures less variance than PC1, it still accounts for a significant amount of the remaining variability. The values of the explained variance values for the first two of the principal components. PC1 for BraTS 2018 and 2019 for LGG and HGG are 0.9179, 0.9066 and 0.9152, 0.9087 respectively. PC2 for BraTS 2018 and 2019 for LGG and HGG are 0.0760, 0.0756 and 0.0771, 0.0742 respectively. It is seen that PC1 gives the highest variance value for both HGG and LGG. By highlighting the original variables, first-order energy for both ROI2 ofHGG and LGG, which contribute the most to this component, PC1, assists with feature selection due to its high loading values. Thus, first-order Energy ROI2 has been considered for further processing.

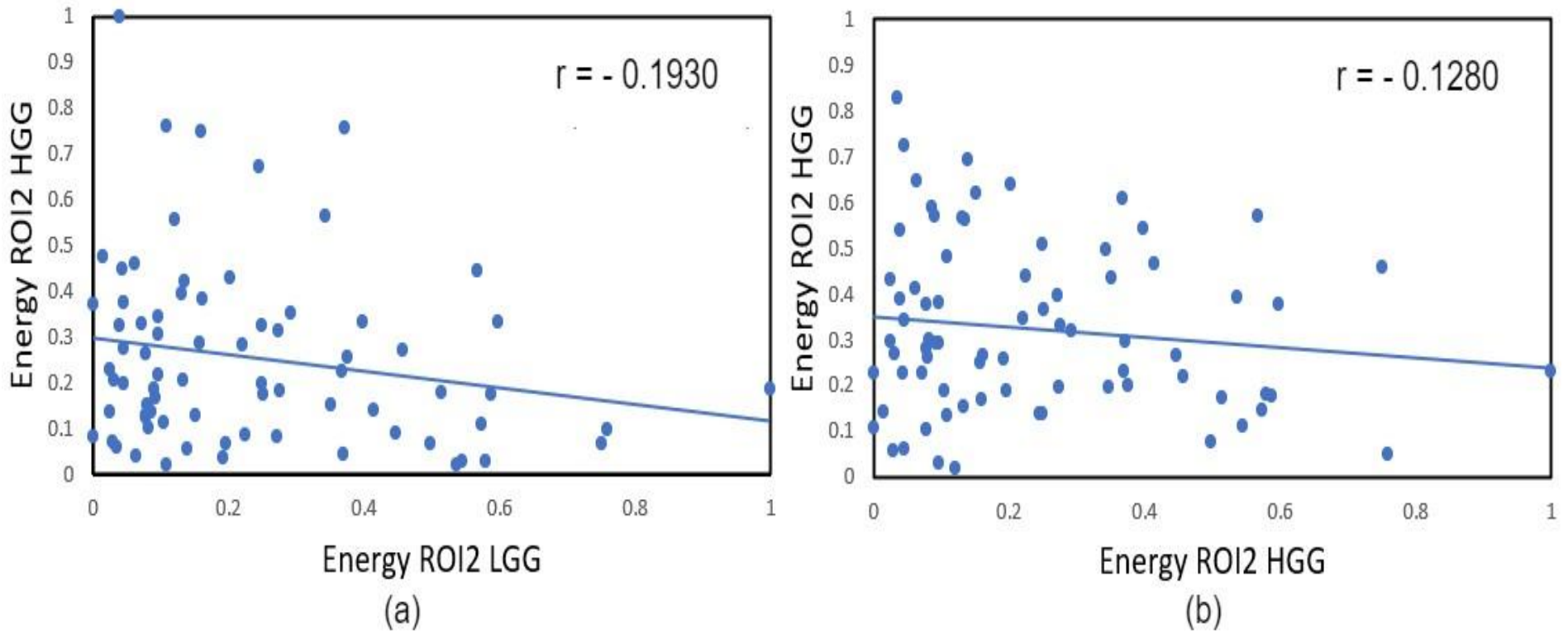

**Figure 8:** Plots representing the correlation (r) energy values of ROI2 LGG and HGG (a) BraTS 2018 dataset and (b) 2019 dataset

Figure 8(a) and 8(b) shows the correlation plots of first significant principal component(first-order Energy of ROI2 denoted by Energy ROI2) obtained from the PCA for LGG and HGG respectively for BraTS 2018 and 2019 datasets. Notably, larger energy values were found in tumors with HGG, which can be correlated to higher necrosis of the tissue [8]. Also, non-homogeneous textures, characterized by their intricate patterns, typically feature more high-frequency details, edges, and variations, lead to higher energy components compared to smoother, homogeneous textures. Thus, in this work, energy feature was considered for the overall tumor grade discrimination as HGGs and LGGs.For BraTS 2018 and 2019, the recorded Pearson's correlation coefficient $r$ for BraTS 2018 and 2019 are 0.1931 and 0.1280 respectively. It is seen that the relationship between the energy values ROI2 of HGG and LGG have very low correlation.This indicates that the data is well separated and is used to distinguish between LGG and HGG.

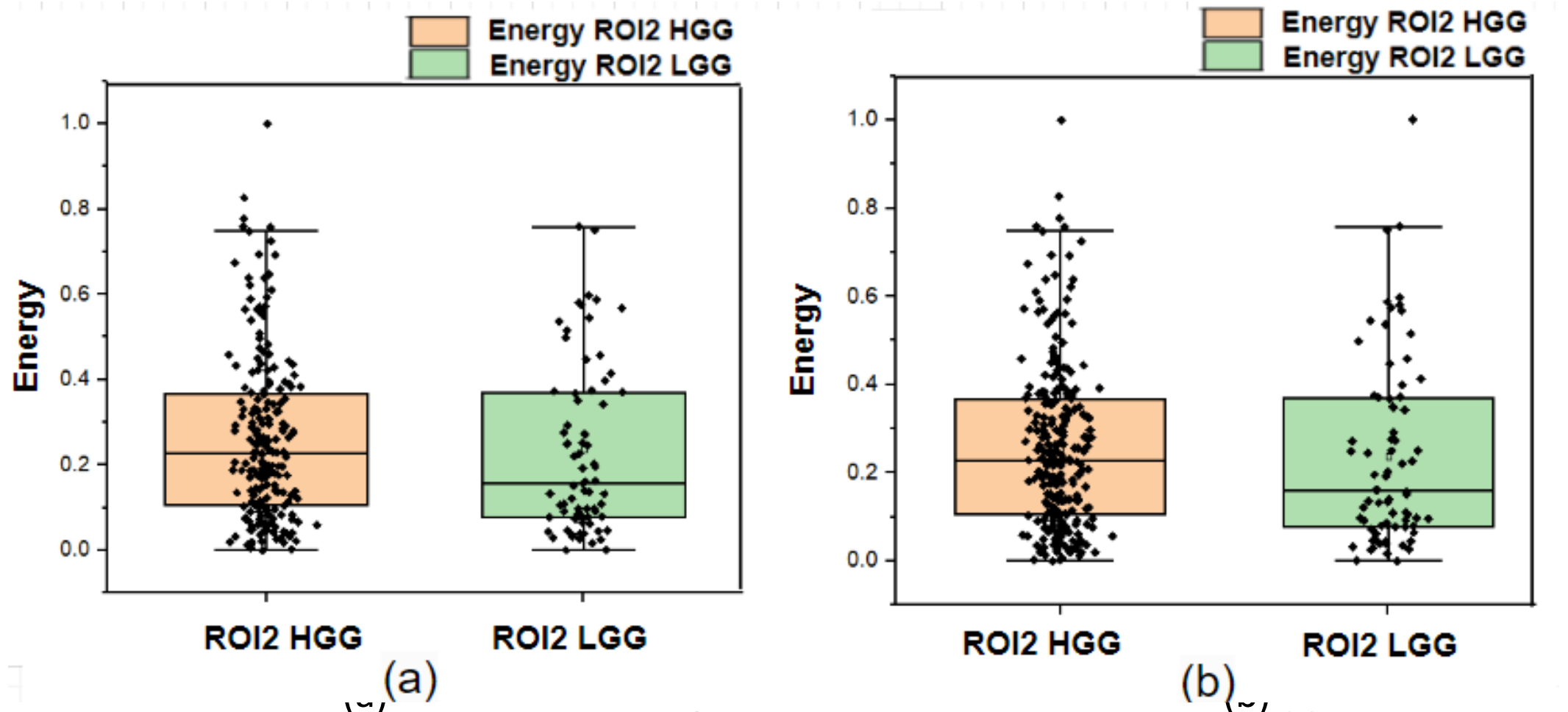

**Figure 9:** Box plots representing the energies of the ROI2 LGG and HGG images for (a) BraTS 2018 and (b) 2019 datasets

Figure 9(a) and (b) show the box plots for BraTS 2018 and 2019 respectively. The box plots comparing the energy of ROI2 between HGG and LGG class reveal that the median energy for HGG is noticeably higher than that of LGG, indicating a significant difference in the central tendency between the two groups. Additionally, the box plot for LGG exhibits a wider interquartile range (IQR) compared to HGG, suggesting greater variability in energy values within the LGG class. This signifies the homogeneity of LGG and non-homogeneity and presence of necrotic region in HGG. While there is one outlier in the LGG class, there are a few data points in the HGG class that fall beyond the whiskers, potentially indicating fewer outliers.For 2018, the mean and median values of ROI2 of HGG images are 0.2621 and 0.2277 respectively. For LGG images, mean and median values of ROI2 are 0.2365 and 0.1570 respectively. The mean and median values of HGG are higher than LGG indicating the non-homogeneity nature of HGG. The IQR for HGG and LGG data are 0.2600 and 0.2919 respectively. BraTS 2018 has a skewness of 1.01 for HGG data and 1.25 for LGG data. The higher skewness of LGG indicates the uniform nature of the LGG. For 2019 dataset, the mean and median values of ROI2 of HGG data are 0.2541 and 0.2262 respectively. For LGG data, mean and median values of ROI2 are 0.2379 and 0.1585 respectively. The higher values of the mean and median illustrates that HGGs had higher frequency of pixels which characterises necrosisas compared to LGGs. The IQR for HGG and LGG data are 0.2608 and 0.2910 respectively. BraTS 2019 has a skewness of 1.01 for HGG data and 1.23 for LGG data indicating that the dataset is right skewed[29].It has been established by the right skewness in the box plots of Figure 9 that energy feature of HGG is high [8]. Thus, the energy values of ROI2 plays a significant role in the analysis of glioma grading.

The efficacy of glioma classification algorithms has been assessed using quantitative classification metrics. The classification metrics for BraTS 2018 and BraTS 2019 are given in Table 4 and Table 5 for training, and five-fold cross-validation, respectively. The performance of XGBoost and Random Forest classifiers during the training resulted in 100% for all the evaluation metrics. However, during the validation of the models, the SVC's metrics are better when compared to the other two classifiers recording an accuracy of 90.17%, precision of 91.04%, recall of 96.19%, F1-score of 93.53% and AUC of 98.46%. Table 7 shows the specificity values for all the classifiers. It can be seen that the fusion algorithm classifies 96.19% of cases with true positives and 3.81% of cases with false negatives for the SVM classifier on BraTS 2018. With 73.33% specificity for SVM, the classifier predicts 73.33% of cases as true negatives and 26.67% as false positives[30]. Similar results are seen on the BraTS 2019 dataset[30].

The confusion matrices for the XGBoost, SVC, and RFC for BraTS 2018 and BraTS 2019 are shown in Figure 10. In this implementation, it can be seen that for the BraTS 2018 dataset, of the total 210 HGG cases, 197 were correctly classified, and 13 cases were misclassified by the XGBoost classifier. Of the 75 cases of LGG, 52 were classified accurately, and 23 were misclassified. It is seen that XGBoost performed better at classifying the glioblastoma cases than the low-grade gliomas. A similar performance was seen when XGBoost was implemented on the BraTS 2019 dataset. When the Random Forest Classifier is considered, a similar analysis can be seen where the number of cases classified correctly for HGG was greater than those classified as LGG. Nevertheless, the Support Vector Classifier performed better than the other two classifiers as the number of cases correctly classified was higher with 202 cases than the misclassified cases.

| AUC | Accuracy | Precision | Recall | F1 Score | AUC |
|---|---|---|---|---|---|
| 2018 | 2019 | | | | |
| 100% | 100% | 100% | 100% | 100% | 100% |
| 98.46% | 94.32% | 94.36% | 98.55% | 96.41% | 98.15% |
| 100% | 100% | 100% | 100% | 100% | 100% |

**Table 4:** Training performance metrics computed for BraTS datasets

| Classifiers | Accuracy | Precision | Recall | F1 Score | AUC | Accuracy |
|---|---|---|---|---|---|---|
| | 2018 | | | | | |
| XGBoost | 100% | 100% | 100% | 100% | 100% | 100% |
| SVM | 94.29% | 94.18% | 98.33% | 96.21% | 98.46% | 94.32% |
| RFC | 100% | 100% | 100% | 100% | 100% | 100% |

**Table 5:** Five-fold cross validation performance metrics computed for BraTS datasets

| Classifiers | Accuracy | Precision | Recall | F1 Score | AUC | Specificity | Accuracy | Precision | Recall | F1 Score | AUC | Specificity |
|---|---|---|---|---|---|---|---|---|---|---|---|---|
| | 2018 | | | | | | 2019 | | | | | |
| XGBoost | 87.36% | 89.64% | 93.80% | 91.62 % | 90.41 % | 69.33% | 88.95% | 89.73% | 96.90% | 93.14 % | 91.73 % | 61.83% |
| SVM | 90.17% | 91.04% | 96.19% | 93.53 % | 94.60 % | 73.33% | 91.34% | 93.05% | 96.13% | 94.53 % | 93.71 % | 74.75% |
| RFC | 88.77% | 91.25% | 93.80% | 92.48 % | 92.92 % | 74.66% | 88.65% | 89.42% | 96.88% | 92.95 % | 89.04 % | 60.41% |

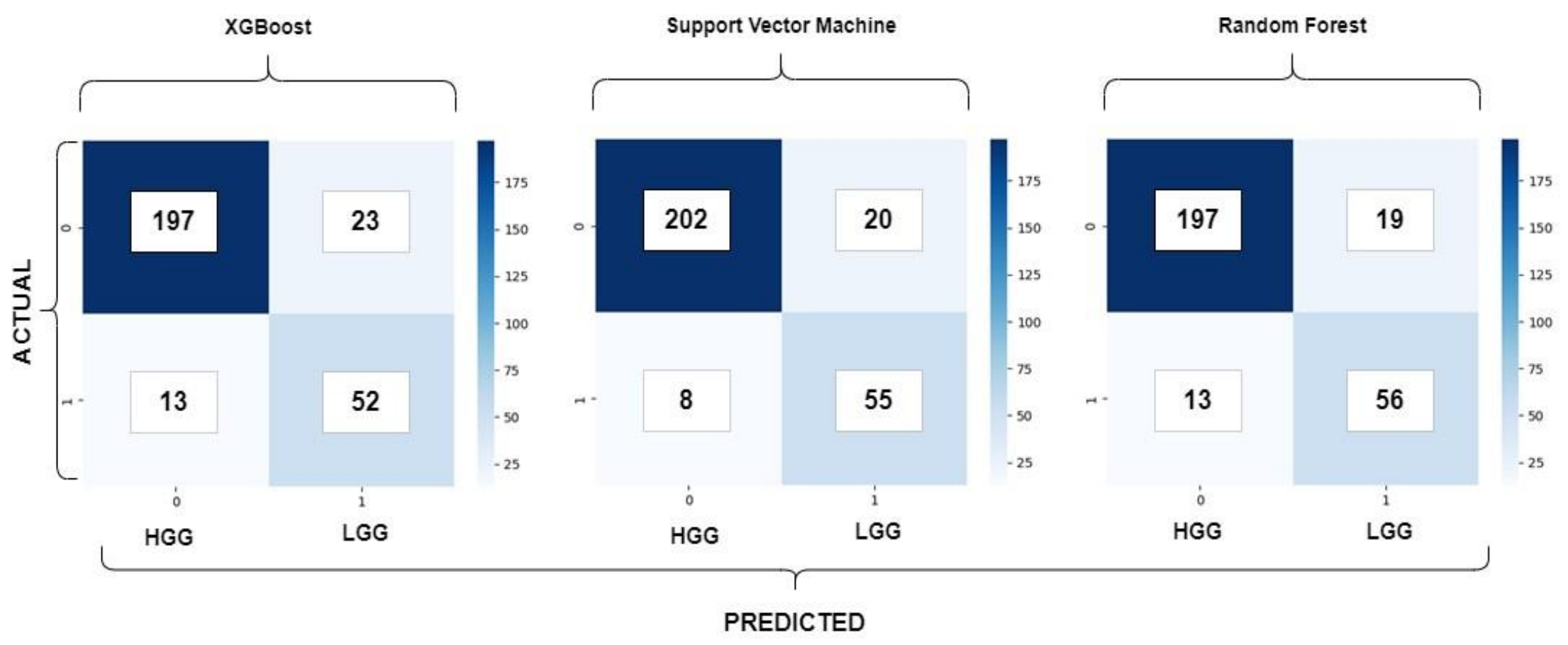

(a)

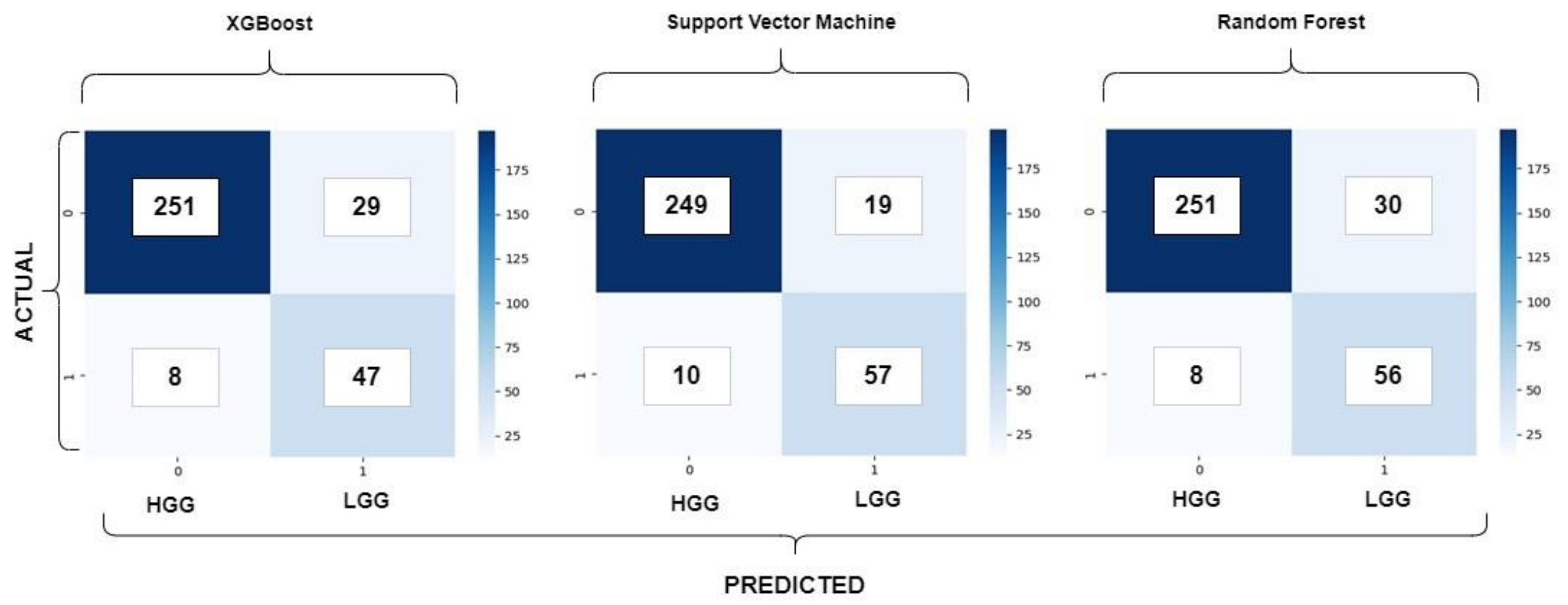

(b)

**Figure 10:** Confusion matrices of XGBoost, Support Vector Machine, and Random Forest classifiers for 5 fold cross validation (a) BraTS 2018 dataset (b) BraTS 2019 dataset

The ROC curve offered a more thorough assessment of the classifiers' performance in the BraTS datasets, which is imbalanced, where the HGG cases are significantly more prevalent than LGG cases. This is because it took distinct trade-offs between the two classes into account. The ROC curves are given in Figure 11 for each BraTS 2018 and BraTS 2019. Figure 11(a) shows the ROC for XGBoost, 11(b) shows the ROC for the Support Vector Machine, and 11(c) shows the ROC for the Random Forest Classifier. Figure 11(a), (b), (c) are implemented on BraTS 2018, and Figure 11(d), (e), (f) are implemented on BraTS 2019.

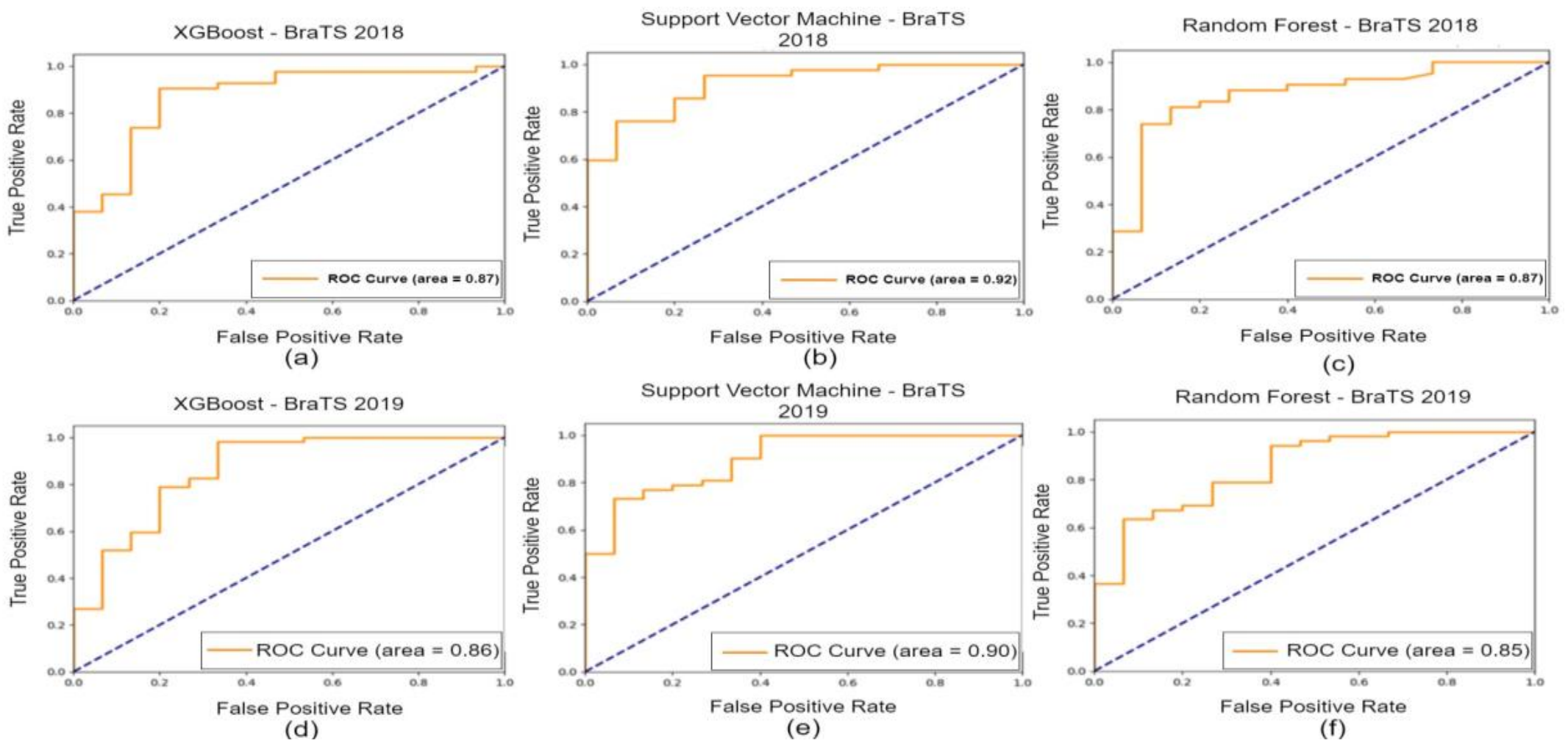

**Figure 11:** ROC plot of classifiers (a) and (d) XG Boost (b) and (e) Support Vector Classifier (c) and (f) Random Forest Classifier for BraTS 2018 and 2019 datasets

Table 6 indicates that the proposed technique outperforms the current methods in literature and proves that fusing the wavelets for high-quality radiomicsfeature extraction improves accuracy. The accuracy obtained for classification is 90.17% and 91.34% for BraTS 2018 and 2019 datasets respectively.

**Table 6:** Comparison of proposed method results with research in the literature

| Research Work | Feature Extraction | Classification Methods | Datasets | Accuracy | Precision | Recall | F1 Score | AUC |
|---|---|---|---|---|---|---|---|---|
| Dequidt et al., 2021 | Radiomics | SVM | BraTS | 84.10% | NA | 87.0% | NA | NA |
| Cui et al., 2019 | LASSO | RF | BraTS 2015 | 91.30% | NA | NA | NA | 95.60% |
| Kumar et al., 2020 | Wavelets | RF | BraTS 2018 | 97.54% | NA | 97.62% | 98.3% | 97.48% |
| Proposed Work | Wavelets, Radiomics with PCA | XGBoost, SVM and RF | BraTS 2018 | 90.17% | 91.00% | 96.10% | 93.53% | 94.60% |
| Proposed Work | Wavelets, Radiomics with PCA | XGBoost, SVM and RF | BraTS 2019 | 91.34% | 93.00% | 96.10% | 94.53% | 93.70% |

## 5. CONCLUSION

This study focused on glioma classification into LGG and HGG, implementing a wavelet-based feature fusion approach. The classification utilized multi-sequence MRI data from the BraTS 2018 and 2019 datasets. The application of fusion and aggregation techniques ensured the integration of crucial information from diverse MRI sequences and segmented tumor regions, thereby enhancing the overall classification process. The decision to utilize radiomics feature extraction was motivated by its efficiency and robustness in capturing textural information from the fused images. The effective use of PCA for dimensionality reduction played a key role in preserving essential information while transforming high-dimensional features into a more manageable lower-dimensional space.

The achieved high accuracy can be attributed to the novelty of the wavelet fusion algorithm. Notably, the SVM classifier exhibited exceptional performance, yielding accuracy, precision, recall, F1 score, and AUC values of 91.34%, 93.00%, 96.10%, 94.53%, and 93.70%, respectively. The potential of this work and future research avenues could explore the extension of this study to the grading of gliomas (I-IV), thereby expanding the scope and applicability of the proposed methodology.